\documentclass{article}
\usepackage{ijcai22}

\usepackage{times}

\usepackage{soul}
\usepackage{url}
\usepackage[utf8]{inputenc}
\usepackage[small]{caption}
\usepackage{graphicx}
\usepackage{amsmath}
\usepackage{booktabs}
\usepackage{amsthm,amsmath,amssymb}
\usepackage[hidelinks]{hyperref}

\usepackage{color}
\usepackage{mathrsfs}
\newtheorem{myDef}{Definition}

\usepackage{bibspacing}
\title{Graph-level Neural Networks: Current Progress and Future Directions}

\author{
Ge Zhang$^\dag$\and
Jia Wu$^\dag$\and
Jian Yang$^\dag$\and
Shan Xue$^{\ddag}$\and
Wenbin Hu$^\S$\and 
Chuan Zhou$^\natural$\and\\
Hao Peng$^\sharp$\and
Quan Z. Sheng$^\dag$\and
Charu Aggarwal$^\pounds$\\
\affiliations
$^\dag$ School of Computing, Macquarie University, Sydney,  Australia\\
$^\ddag$ School of Computing and Information Technology, University of Wollongong, Australia\\
$^\S$ School of Computer Science, Wuhan University, Wuhan, China \\
$^\natural$ Academy of Mathematics and Systems Science,
Chinese Academy of Sciences, Beijing, China\\
$^\sharp$ School of Cyber Science and Technology, Beihang University, China\\
$^\pounds$ IBM T. J. Watson Research Center
Yorktown Heights, New York, USA \\
\emails
\{ge.zhang@hdr., jia.wu@, jian.yang@, michael.sheng\}mq.edu.au, sxue@uow.edu.au, hwb@whu.edu.cn, zhouchuan@amss.ac.cn, penghao@buaa.edu.cn, charu@us.ibm.com
}

\begin{document}

\maketitle

\begin{abstract}
Graph-structured data consisting of objects ({\it i.e.}, nodes) and relationships among objects ({\it i.e.}, edges) are ubiquitous. Graph-level learning is a matter of studying a collection of graphs instead of a single graph. Traditional graph-level learning methods used to be the mainstream. However, with the increasing scale and complexity of graphs, Graph-level Neural Networks (GLNNs, deep learning-based graph-level learning methods) have been attractive due to their superiority in modeling high-dimensional data. Thus, a survey on GLNNs is necessary. To frame this survey, we propose a systematic taxonomy covering GLNNs upon deep neural networks, graph neural networks, and graph pooling. The representative and state-of-the-art models in each category are focused on this survey. We also investigate the reproducibility, benchmarks, and new graph datasets of GLNNs. Finally, we conclude future directions to further push forward GLNNs. 
The repository of this survey is available at \textcolor[rgb]{0.3,0.2,0.7}{ \href{https://github.com/GeZhangMQ/Awesome-Graph-level-Neural-Networks}{https://github.com/GeZhangMQ/Awesome-Graph-level-Neural-Networks}}.
\end{abstract}

\section{Introduction}

The research on graph-structured data can be traced back to the 18th century \cite{biggs1986graph}. Since then, graph-level learning that takes a number of graphs (\textit{i.e.}, a graph dataset) as the study object has attracted great attention. Graph Isomorphism (GI) that determines whether two graphs are isomorphic\footnote{$\mathcal G_1$ and $\mathcal G_2$ are isomorphic if there exists a bijection function $f$: $\mathcal V(\mathcal G_1)\rightarrow \mathcal V(\mathcal G_2)$, $\forall{u,v} \in \mathcal V(\mathcal G_1)$, $u$, $v$ are connected, \textit{iff} $f(u)$ and $f(v)$ are connected in $\mathcal G_2$, where $\mathcal V$ denotes the node set.} has been the pioneer topic of graph-level learning as far back as the 1940s \cite{bondy1977graph}. In the 1970s, GI was defined as the ``disease" of graph theory, and a lot of algorithms researching it have emerged \cite{weisfeiler1968reduction,mckay2014practical}. Moving into the 21st century, due to the thriving of machine learning and deep learning,  graph classification \cite{shervashidze2011weisfeiler,xu2018powerful}, embedding \cite{narayanan2017graph2vec}, regression \cite{gilmer2017neural}, and generation \cite{you2018graphrnn}) have been the main tasks of graph-level learning.

Graph structures give the unified modeling manner for many real-world entities. For example, the chemical compound, interactions between proteins, and a human brain can be described as the molecular graph, protein-to-protein graph, and brain network, respectively. Graph-level learning has been of great practical significance. For instance, since the 1960s, scientists have utilized graphs to model molecules \cite{morgan1965generation} and designed molecular fingerprints to assist chemical dataset substructure searching \cite{ralaivola2005graph,duvenaud2015convolutional}. Graph regression has been becoming an effective manner for molecular property prediction \cite{gilmer2017neural} and drug discovery \cite{chen2018rise}. Graph classification can discriminate the brains affected by neurological disorders from healthy individuals \cite{wang2017structural}. Most recently, graph-level learning has been applied to the \textit{open catalyst project} that aims to find new catalyst molecules \cite{chanussot2021open}.


\begin{figure}[!t]
	\includegraphics[width=1\linewidth]{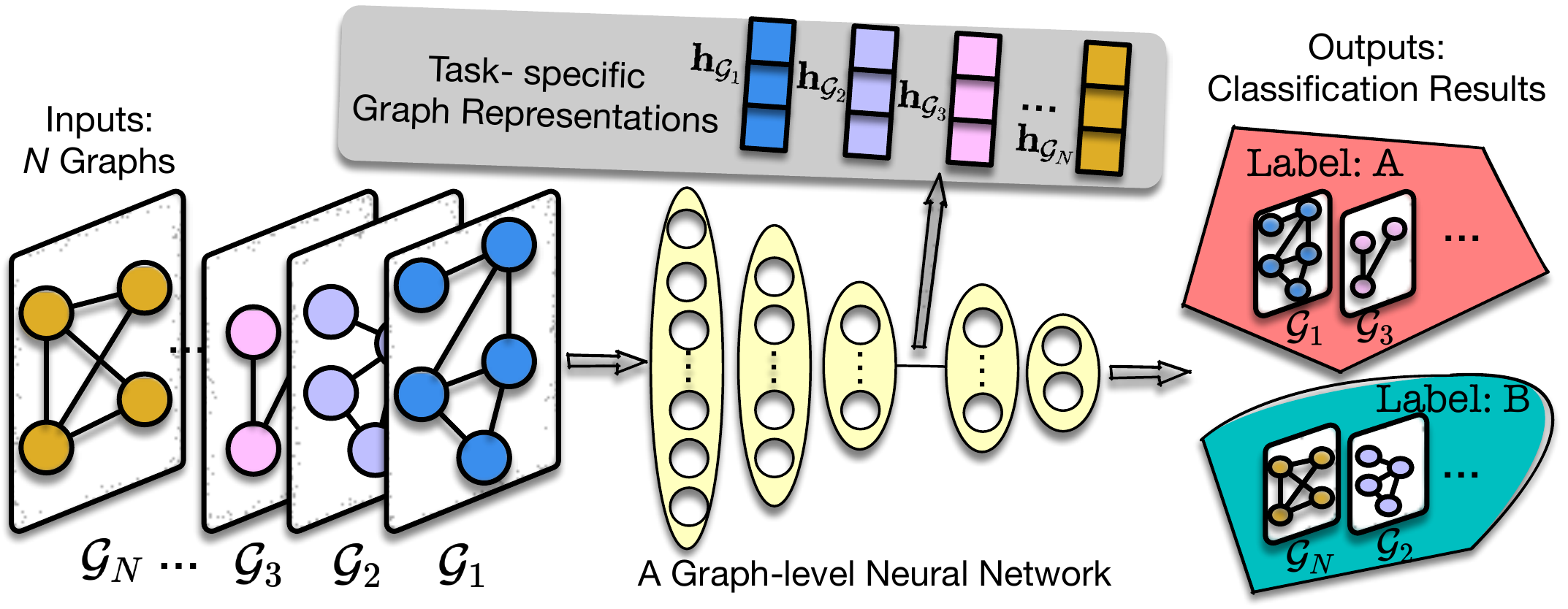}
	\centering
	\caption{A toy example about the representative graph-level classification based on GLNNs. After inputting a graph dataset consisting of $N$ graphs, the GLNN learns task-specific graph representations and outputs classification results of each graph.
	}
	\label{fig1}
	\vspace{-3mm}
\end{figure}

Traditional graph-level learning methods have demonstrated good performance on small-sized graphs but output highly-sparse results for large-scaled graph datasets. In addition, the GI issue that only has quasipolynomial-time solutions \cite{babai2016graph} is involved in some traditional methods. The computational complexity of these methods has quadratic or even exponential growth with the increase of data size. Beyond traditional methods, GLNNs have achieved state-of-the-art performance on graph-level learning, \textemdash particularly when it comes to high-dimensional graph data. GLNNs cater to graph-level learning by: (1) automatically learning nonlinear features from graphs; 2) explicitly encoding graphs into compact numerical representations; 3) learning task-specific graph representations and addressing graph-level learning tasks in an end-to-end fashion (see Figure \ref{fig1}).



\begin{figure*}[!t]
	\vspace{-3mm}
	\includegraphics[width=0.98\linewidth]{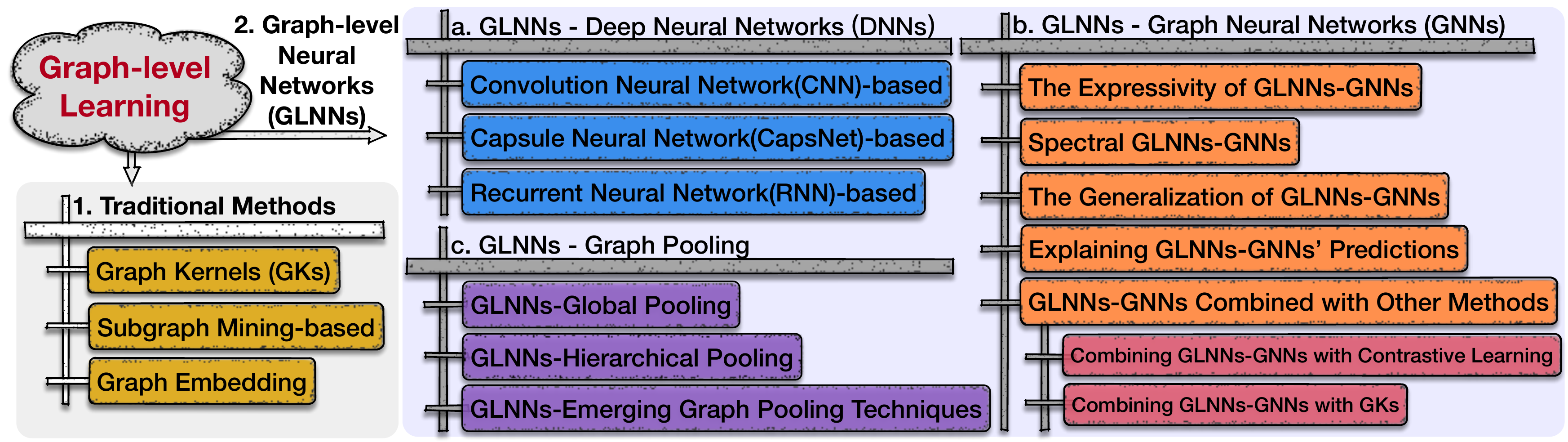}
	\centering
	\caption{The taxonomy of traditional graph-level learning methods and graph-level neural networks.}
	\label{fig2}
	\vspace{-3mm}
\end{figure*}

Existing survey papers mainly concentrating on node-level learning tasks that analyze node/edge/structure of a single graph, such as network embedding \cite{cui2018survey}, community detection \cite{su2021comprehensive}, and anomaly detection \cite{ma2021comprehensive}. There are also two up-to-date surveys \cite{wu2020comprehensive,zhang2020deep} that focus on node-level neural networks. \cite{kriege2020survey} and \cite{nikolentzos2021graph} review graph kernels (one type of traditional graph classification methods), but they only introduce GLNNs roughly and even do not provide taxonomy.



To the best of our knowledge, this is the first survey devoted to GLNNs. This paper is intended to help new practitioners understand graph-level learning and enable researchers to have a grasp of the current progress of GLNNs. The contributions of this survey are summarized as:
\begin{itemize}
\item{\textbf {Systematic Taxonomy:}} We propose a systematic taxonomy for GLNNs (see Figure \ref{fig2}). In each category, representative algorithms and the latest progress are introduced. We also summarized papers about the reproducibility, benchmarks, new datasets of GLNNs.
\item{\textbf {High-impact References:}} Our references mainly come from the well-established conferences in machine learning (\textit{e.g.}, \textit{NeurIPS}, \textit{ICLR}, \textit{ICML}, \textit{KDD}, \textit{AAAI}, \textit{IJCAI}, \textit{CIKM}, \textit{ICDM}, and \textit{WWW}).
\item{\textbf {Future Directions:}} We discuss the future directions of GLNNs and hope this part can inspire researchers and further promote the development of GLNNs-GNNs.
\end{itemize}



\section {Definitions}

\begin{myDef}
\textit{(Graph)}. A graph can be represented as $\mathcal G=\{\mathcal V, \mathcal E \}$, where $\mathcal V$ and $\mathcal E$ denote the set of $n$ nodes and $m$ edges, respectively. The graph structure can be represented by an adjacency matrix 
$\mathbf A \in\{0,1\}^ {n\times n }$, where $\mathbf A_{u,v}=1$ denoted there is an edge between nodes $u$ and $v$, otherwise 0.
The node $v$ and edge $\mathcal E_{u,v}$ can be annotated with vector $\mathbf x_v\in \mathbf X$ and  $\mathbf s_{u,v}\in \mathbf S$, respectively, where $\mathbf X\in \mathbb R^{n \times b}$ and $\mathbf S\in \mathbb R^{m \times d}$.
\end{myDef}

\begin{myDef}
\textit{(Graph-level Learning)} It takes a graph dataset $\mathbb G = \{\mathcal G_1,..., \mathcal G_N\}$ as inputs, and it outputs results of a graph-level learning task. The prominent property of a graph is \textbf{permutation-invariant} \textemdash enumerating the ordering of nodes does not change its structure. Hence, the graph-level learning algorithm is supposed to be permutation-invariant, \textit{i.e.}, the prediction result about a graph does not affect by permutations on its node ordering.
\end{myDef}

\begin{myDef}
\textit{(Graph-level Neural Networks)} GLNNs perform graph-level learning tasks on the graph dataset $\mathbb G$ in an end-to-end fashion. Generally, GLNNs encode the graph $\mathcal G_i\in\mathbb G$ into a vectorized representation (\textit{a.k.a.}, embedding) $\mathbf h_{\mathcal G_i}$, followed that the task-specific results will be output.
\end{myDef}


\section{A Development of Graph-level Learning}
In this section, we briefly review three types of traditional graph-level learning methods first. Then, we explain why the deep learning-based method GLNNs is more desirable. The taxonomy of graph-level learning from traditional methods to GLNNs is shown in Figure \ref{fig2}.

\noindent\textbf{Graph Kernels (GKs).} GKs calculate pairwise kernel values (\textit{i.e.,} similarity) of graphs through well-defined kernel functions to obtain the \textit{Gram matrix} where each entry denotes the kernel value of pairwise graphs. Afterward, employing off-the-shelf machine learning techniques to perform graph-level learning tasks on the \textit{Gram matrix}, such as Support Vector Machines for classification, \textit{K}-Means for clustering, and Gaussian Processes for regression. The representative method include Weisfeiler-Lehman GKs \cite{shervashidze2011weisfeiler}\cite{togninalli2019wasserstein}, graphlet GKs \cite{shervashidze2009efficient}, and random walk GKs \cite{kang2012fast}.

\noindent\textbf{Subgraph Mining-based Approaches.} Frequent subgraph-based methods \cite{thoma2010discriminative} generally represent each graph by a series of frequent subgraphs (\textit{i.e.}, the subgraphs with the occurrence in the graph dataset beyond a threshold) beforehand. Then, machine learning techniques can be applied to handcrafted graph representations. There are also informative subgraph-based \cite{kudo2004application} and dense subgraph-based methods \cite{lanciano2020explainable}.  


\noindent\textbf{Graph Embedding.} Based on handcrafted features, GKs and subgraph mining-based approaches output deterministic instead of learnable results. Moreover, pairwise similarity calculation in GKs and the GI issue associated with subgraph mining-based approaches make these algorithms suffer from computation bottlenecks. Practitioners have turned to data-driven graph embedding, such as graph2vec \cite{narayanan2017graph2vec} and AWE \cite{ivanov2018anonymous}, which can learn knowledge from graphs automatically and output learned task-agnostic graph representations explicitly.

\noindent\textbf{Why GLNNs are Needed?} GLNNs bring significant benefits to the graph-level learning community. They provide a general framework to encode graphs into task-specific representations and predict graphs in an end-to-end system. Benefiting from deep learning, GLNNs can learn the highly nonlinear features from the large-scale sparse graph data effectively. Moreover, non-structural features (\textit{e.g.}, node and edge attributes) can be naturally integrated into GLNNs as the supplements for graph structures. Next, we will discuss GLNNs based on deep neural networks, graph neural networks, and graph pooling in detail.

\section{GLNNs - Deep Neural Networks (DNNs)}
In this section, we survey GLNNs based on three types of DNNs, \textit{i.e.}, Convolution Neural Networks (CNNs), Capsule Neural Networks (CapsNets), and Recurrent Neural Networks (RNNs). These DNNs were originally proposed for learning grid-structured data (\textit{e.g.}, image) and sequence-structured (\textit{e.g.}, text and speech) data, instead of the non-Euclidean structure data \textemdash graphs. 

\subsection{CNN-based Approaches}
Generalizing CNNs to graph-structured data in the high-dimensional and irregular domain is non-trivial, due to the unordered nodes and the unbounded node neighborhood. One of the straightforward ways is to transform each graph to be the girded data. For example, through selecting nodes sequence and assembling regular node neighborhoods, PATCHY-SAN \cite{niepert2016learning} successfully applied CNNs to the graph classification task. On the contrary, some algorithms apply the idea of spatial CNNs to graph-level learning. For example, in DCNN \cite{atwood2016diffusion}, the node feature transformation is based on a three-dimensional tensor that contains the multi-hop node neighborhood information. Through averaging the learned node representations of each graph, DCNN obtains the graph representation. Another method ECC \cite{simonovsky2017dynamic} learns the node representation by applying weighted summation to its neighbors' representations coming from the former neural layer. The weights are edge-specific and learned by the feature transformation on edge attributes. ECC coarsens a graph into a node to obtain the graph representation.


\subsection{CapsNet-based Approaches }
CapsNets \cite{hinton2011transforming} were proposed to address the issue that CNNs cannot effectively capture spatial relationships among features. CapsNets extend the scalar neuron to a vector-structured capsule, in which each vector element is responsible for capturing different features such as position, pose, and texture of the input. Analogically, GCAPS-CNN \cite{verma2018graph} and GapsGNN \cite{xinyi2018capsule} leverages the idea of CapsNets to preserve high-level properties of graphs into graph representations. Moreover, GCAPS-CNN achieves permutation-invariance by calculating the covariance of layer outputs.


\subsection{RNN-based Approaches}
As a type of neural networks in which all nodes are connected directly or undirectly, RNNs are specialized in capturing sequential patterns in data \cite{medsker2001recurrent}. \cite{you2018graphrnn} proposed an RNN-based graph generation model in which nodes and edges are formed in a sequence. \cite{lee2018graph} emphasized that significant structures are confined to a small part of a graph. Hence, they used substructures sampled by RNNs to do the graph classification task. 

\section{GLNNs - Graph Neural Networks (GNNs)}
Currently, GNNs are the standard toolkit for learning node representations \cite{wu2020comprehensive}.
As the key component of GNNs, graph convolution can be defined either in the spatial domain or spectral domain. To achieve permutation-invariant GLNNs-GNNs for graph-level learning, a trivial idea is to employ a GNN followed by a readout function, such as summation-based, averaging-based, and jumping knowledge-based functions. The former two types of readout obtain the graph representation $\mathbf h_\mathcal G$ by summing/averaging node representations $\mathbf h_v, \forall v\in{\mathcal V (\mathcal G)}$. The latter one aggregates node representation across all convolution layers, that is $\mathbf h_G = \sum_{v\in \mathcal V(\mathcal G)} \left[\mathbf h_v^{(0)}; ...; \mathbf h_v^{(L)}\right]$ where $\mathbf h_v^{(l)}$ is the representation of node $v$ output by the $l$-$th$ graph convolution layer, and ``$;$" denotes the concatenation operation. The most widely used GLNNs-GNNs are Messaging Passing Neural Networks (MPNNs) \cite{gilmer2017neural}. MPNNs belong to spatial graph convolution where each node updates its representation by aggregating vectorized representations of its neighbors.


We investigate the expressivity of GLNNs-GNNs first. Then, spectral GLNNs-GNNs will be discussed. In addition, we introduce models focusing on improving the generalization of GLNNs-GNNs and explaining GLNNs-GNNs' predictions. This section ends with a discussion of employing other techniques to improve GLNNs-GNNs.

\subsection{The Expressivity of GLNNs-GNNs}
Since GLNNs-GNNs are proposed for graph-level learning, it is critical to investigate the power of GLNNs-GNNs in distinguishing graph structures. Generally, researchers call such a power ``expressivity". Employing the Weisfeiler-Leman graph isomorphism testing (WL-test, \cite{weisfeiler1968reduction}) to describe GLNNs-GNNs' expressivity is attractive at present. The WL-test contains the neighborhood aggregation-based message passing similar to MPNNs. \cite{xu2018powerful} firstly proved that the upper bound of MPNNs in distinguishing graph structures is the 1-dimensional WL test (1-WL). They emphasized that it is the injective message passing making 1-WL so powerful. By adding multi-layer perceptrons behind each graph convolution layer to guarantee the injective message passing, GIN that is equivalent to the 1-WL is proposed. However, 1-WL equivalence MPNNs cannot distinguish graphs that are indistinguishable by 1-WL, such as regular graphs. Also, 1-WL cannot count substructures in a graph other than the star-shaped pattern \cite{arvind2020weisfeiler,chen2020can}, but substructures (\textit {e.g.}, triangles, cycles) are critical for learning social networks and molecular graphs.

By introducing more inductive biases, scientists have presented MPNNs beyond the expressivity of 1-WL. For example, in analogy to \textit{k}-WL (the extension of 1-WL), \cite{morris2019weisfeiler} proposed \textit{k}-GNNs which perform message passing on pre-defined subgraphs consisting of $k$-nodes and achieves the same expressivity as \textit{k}-WL. However, when $k \ge 3$, achieving \textit{k}-GNNs is at the cost of exponential increase in computing complexity, since \textit{k}-GNNs perform calculations on \textit{k}-ranked tensors. \cite{maron2019provably} proposed PPGN, which achieves the 3-WL equivalent expressivity by designing a 2-GNN augmented with a quadratic operation. In addition, \cite{bouritsas2020improving} and \cite{bevilacqua2022equivariant} enriched MPNNs through employing pre-processing substructures as additional features and automatically selecting discriminative substructures, respectively.

Practically, theoretically more powerful MPNNs fail to beat their 1-WL equivalent counterparts on most graph datasets \cite{dwivedi2020benchmarkgnns}. This is because when graph attributes can perform as supplements of structural information, almost all graphs can be distinguished by 1-WL. Hence, it is not worth achieving more powerful MPNNs at the cost of high complexity. Adding unique node identities \cite{murphy2019relational} or random features \cite{sato2020survey} into graphs can achieve more powerful MPNNs within linear complexity. In addition, breaking the key property of graph convolution - local updates \cite{battaglia2018relational} is the other reason why theoretically more powerful GLNNs-GNNs do not achieve superior performance in experiments. To tackle this problem, \cite{balcilar2021breaking} proposed GNNML3 based on the eigendecomposition of graph Laplacian in the spectral domain. Through encoding graph signals from various frequencies, and casting critical computation operations in Matrix Query Languages (MATLANG, that is originally proposed to measure the expressivity of linear algebra) into graph convolution, GNNML3 is equivalent to 3-WL experimentally. \cite{bodnar2021weisfeiler} introduced the simplicial complexes into graph convolution to achieve GLNNs-GNNs with expressivity no less than 3-WL while ensuring local updates. Notably, inspired by MATLANG, \cite{geerts2022expressiveness} presented that describing GLNNs-GNNs' expressivity can be released from the intricacies of the WL-test by employing tensor languages to model GLNNs-GNNs.


\subsection{Spectral GLNNs-GNNs}
Existing MPNNs work as a low-pass filter that can only preserve graph signals in low frequencies \cite{balcilar2020analyzing}. However, multi-scale graph signals are critical for graph-level learning, say, graph signals in high frequencies can highlight the differences among graphs. Spectral GLNNs-GNNs that define graph convolution in the spectral domain can capture multi-scale graph signals under robust theoretical guarantees. In addition to GNNML3, other spectral GLNNs-GNNs also achieve good performance on graph-level learning. For example, \cite{zheng2021framelets} defined framelet-based graph convolution that can filter high- and low-frequency signals by tensorized framelet transforms. Based on the spectral density computation, \cite{sawlani2021fast} proposed a method to capture graph signals in multi-scale beyond the extremum. After performing the spectral analysis about mainstream GNNs, \cite{balcilar2020analyzing} presented a spectral graph convolution with a custom frequency profile that achieves good performance on graph classification.


\subsection{The Generalization of GLNNs-GNNs}

Investigating the generalization ability of GLNNs-GNNs from small to large graphs is necessary since real-world graphs have various sizes and labeling big graphs is difficult. \cite{xu2020neural} theoretically analyzed the size-generalization capability of GLNNs-GNNs that have a single graph convolution layer.
\cite{yehudai2021local} suggested detecting local structures that are both significant in small-sized and large-sized graphs to improve GLNN-GNNs in size generalization. On the other hand, \cite{ma2020adaptive} and \cite{chauhan2020few} explored how to improve the generalization of GLNNs-GNNs under the few-shot setting where only a few (even zero) graphs have the same graph label with unseen graphs. In addition, \cite{alon2021on} emphasized that most of GLNNs-GNNs have the over-squashing issue that each node's receptive field grows exponentially, but long-range interactions among nodes cannot be captured. However, long-range interactions are critical, say, the property of methylnonane compound depends on the atoms located on the molecule's opposite sides. Empirically, appending a fully-adjacent layer in which nodes are fully-connected can help GLNNs-GNNs capture long-range interactions.
 
\subsection{Explaining GLNNs-GNNs' Predictions}

The black-box nature of GLNNs prevents academia and industries from trusting their predictions. The mainstream to explain GLNNs-GNNs is to detect nodes and substructures in a graph that can dominate GLNNs-GNNs' predictions on it. These explanation models work in a post-hoc fashion and can be roughly divided into perturbation-based \cite{yuan2020xgnn,luo2020parameterized,ying2019gnnexplainer}, surrogate model-based \cite{huang2020graphlime}, and gradient-based \cite{pope2019explainability}. One drawback of these models \cite{ying2019gnnexplainer,yuan2020explainability} is that they can only perform on instance-level \textemdash a complete training of the explanation model can only explain the prediction of GLNNs-GNNs on a single graph. \cite{yuan2020xgnn} proposed the reinforcement learning-based XGNN that can explain GLNNs-GNNs' explanation on a specific graph class after a complete training. Furthermore, \cite{luo2020parameterized} presented PGExplainer to explain GLNNs-GNNs collectively and inductively.

\subsection{GLNNs-GNNs Combined with Other Methods}
In this subsection, we discuss algorithms improving GLNNs-GNNs by combining them with contrastive learning and GKs:

\noindent\textbf{Combining GLNNs-GNNs with Contrastive Learning:} Most GLNNs-GNNs work in a supervised fashion. However, annotating graphs is costly. For example, labeling chemical compounds need Density Functional Theory calculation. It leads to the demand for self-supervised techniques. Contrastive learning \cite{hjelm2019learning} has attracted a surge of interest. Generally, contrastive learning-based models train an encoder by contrasting representations having statistical dependencies and those that do not. Analogically, \cite{sun2019infograph} proposed InfoGraph that learns task-agnostic graph representations by maximizing statistical dependencies between the graph and its nodes and minimizing statistical dependencies between the graph and nodes from other graphs. \cite{you2020graph} investigated the impact of data augmentation in graph contrastive learning. \cite{qiu2020gcc} proposed contrastive learning-based graph presentation pre-training model that is optimized by maximizing the agreement between graphs' original and augmented views.

\noindent\textbf{Combining GLNNs-GNNs with GKs:} Leveraging the benefits of GKs to improve GLNN-GNNs is feasible \cite{lei2017deriving} \cite{chen2020convolutional}. For example, GLNNs-GNNs are hard to train due to non-convex objective functions and over-parameterized properties. \cite{du2019graph} proposed a computationally efficient graph classification model in which the graph convolution is replaced by the calculation of pairwise graph kernel values. \cite{long2021theoretically} noticed that GKs naturally involve the substructure comparison. Hence, they introduced random walk-based GKs into graph convolution to improve GLNNs-GNNs in terms of distinguishing graphs.

\section{GLNNs: Graph Pooling}

Graph pooling performs graph-level learning by downsizing the graph. However, it is non-trivial due to the irregular structure of graphs. Recently, differentiable graph pooling has been the mainstream. It can be roughly divided into two types of approaches, \textit{i.e.}, the global and the hierarchical pooling. 

\subsection{Global Pooling}
Global pooling, acting as a bridge between the graph representation and output layers, is generally utilized only once in a GLNN. The simplest global pooling is the readout function. A number of GLNNs \cite{xu2018powerful} \cite{atwood2016diffusion} \cite{simonovsky2017dynamic} apply the readout function on nodes to obtain graph representations. DGCNN \cite{zhang2018end} is a graph classification model leveraging the sorting-based global pooling. After encoding nodes by graph convolution similar to GCN \cite{kipf2016semi}, nodes in each graph will be sorted based on their structural roles and graphs can be pooled into the same size. Afterward, CNNs can be employed to classify graphs. Compared with PATHY-SAN \cite{niepert2016learning} that also transforms graphs to be girded data, DGCNN can be optimized by back-propagation since it can record the sorting order of nodes. \cite{lee2021learnable} mapped nodes to the latent space composed of $k$ structural prototypes and pooled nodes in the same prototype together. The graph representation can be obtained by concatenating embeddings coming from different prototypes.

\subsection{Hierarchical Pooling}
Hierarchical pooling learns coarsen-grained graph structures by gradually applying the pooling layer to downsize the graph. One type of hierarchical pooling is the \textit{cluster assignment-based} method that clusters nodes in a graph and regards clusters as new nodes to form the pool graph. After several pooling operations, the graph can be coarsened into a node and represented by a vector. The cluster assignment can be obtained via deterministic \cite{bruna2014spectral} and learnable clustering. DiffPool \cite{ying2018hierarchical} is one of the representatives based on learnable clustering. Compared with inherently flat GLNNs-GNNs, DiffPool can learn hierarchical representations for graphs. Analogy to DiffPool, other cluster-assignment-based graph pooling methods, such as MinCutPool \cite{bianchi2020spectral}, MemGNN  \cite{Khasahmadi2020Memory-Based} and SAGPool \cite{lee2019self} are proposed. In addition, Top-K Pooling \cite{gao2019graph} is the \textit{node selection-based} hierarchical pooling that selects the nodes with top-$k$ scores into the pooled graph in each pooling layer. In DiffPool, storing dense cluster assignment matrices needs quadratic storage complexity. \cite{cangea2018towards} proposed to use the top-$k$ node selection to achieve hierarchical pooling without sacrificing sparsity. 

Some practitioners proposed that the rich structural information is overlooked by existing graph pooling. They tackled the problem by encoding high-order structures explicitly. For example, \cite{ma2019graph} and \cite{yuan2020StructPool} introduced the graph Fourier transform and conditional random fields into graph pooling to capture more structural information, respectively. \cite{ranjan2020asap} scored substructures instead of nodes for top-$k$ selection.


\subsection {Emerging Graph Pooling Techniques}

Pooling layers are usually put into GLNNs-GNNs as the component to downsize graphs. \cite{mesquita2020rethinking} conducted a sanity check for investigating the impact of hierarchical pooling on GLNNs-GNNs. Experiment results demonstrate that the cluster assignment-based hierarchical pooling does not witness a performance decrease when adopting random clustering or performing clustering on the complement graph. A possible reason is that graph convolution preceded by pooling can smooth graphs sufficiently. Hence, conducting more ablation experiments to validate newly proposed hierarchical pooling methods is necessary. In addition, existing graph pooling fails to keep stable performance on different datasets. \cite{wei2021pooling} proposed to address this problem by the data-specific graph pooling that can adaptively sample off-the-shelf pooling methods for the input graph dataset.


\section{Reproducibility, Benchmarks, and Datasets}
Lacking standard experiment procedures has been a major threat to the reproducibility of GLNNs. \cite{errica2019fair} re-evaluated  GLNNs-GNNs on the graph classification task under rigorous experiments concerning unified train-validation-test split, model selection and assessment processes. Furthermore, they also provided two graph structure-agnostic baselines to help practitioners characterize the contribution of newly proposed GLNNs-GNNs. \cite{dwivedi2020benchmarkgnns} are also concerned by the reproducibility issue. Based on PyTorch and DGL \cite{wang2019dgl}, they presented a benchmarking framework that fairly evaluates GLNNs-GNNs on graph classification and regression tasks.

Existing graph datasets (\textit{e.g.}, TU collection \cite{morris2020tudataset}) are criticized for their small size, isomorphism bias \cite{ivanov2019understanding}, and random data splits. Recently, practitioners have presented new graph datasets to improve the graph-level learning community. For example, \cite{freitas2020large} introduced a graph dataset named MalNet for graph-level learning. \cite{hu2020open} proposed the Open Graph Benchmark (OGB), in which graph datasets are large-scaled, have multiple task categories, and are from diverse domains. Notably, these new datasets propose challenges for GLNNs in scalability, imbalanced learning, explainability, and out-of-distribution generalization.

\section{Future Directions}

In this section, we highlight seven future directions for GLNNs based on the latest progress. 

\subsection{Equivariant GLNNs-GNNs}
It is the permutation-invariant property of message passing that limits GLNNs-GNNs' expressivity \cite{kondor2018covariant}\cite{wijesinghe2022a}.  \cite{de2020natural} proposed permutation-equivariant message passing \textemdash the local output is changed corresponding to the permutation. In addition, practitioners proposed message-passing mechanisms that are equivariant to more graph transformation. For example, \cite{klicpera2021gemnet} presented GemNet in which the message passing is graph permutation- and rotation-equivariant. \cite{satorras2021n} introduced graph translation and reflection -equivariant message passing into GLNNs-GNNs. The above two methods achieve state-of-the-art performance on molecular property prediction. Designing graph transformation-equivariant message passing to improve GLNNs-GNNs' expressivity is a new trend.

\subsection{Graph Pooling with Difference Awareness}
Existing graph pooling especially the top-$k$ node selection-based has argued that nodes with great importance should be placed into pooled graphs. However, there is no rigorous definition of ``importance". In practice, graph convolution acts as a low-pass filter, and nodes with high similarities are put into pooled graphs \cite{mesquita2020rethinking}. However, a pooled graph consisting of dissimilar nodes is critical for graph-level learning. For example, a pair of different atoms can empower completely different properties to two molecules. How to construct a node difference-aware graph pooling method has not been explored yet.

\subsection{Imbalanced Graph-level Learning}
Learning imbalanced data in which a significant gap among the number of instances in different classes is a longstanding issue faced by machine learning. When learning the data with imbalanced class distributions, algorithms usually demonstrate poor generalization in the class with minority instances. However, almost all GLNNs overlook this issue. \cite{freitas2020large} offered a new graph dataset that has imbalanced class distribution. The graph-level learning community urgently needs the baselines targeting imbalanced learning.

\subsection{Out-of-Distribution Generalization of GLNNs}
GLNNs assume that train and test data from the same distribution. However, the assumption is usually invalid in the open-world environment. Testing data could be out-of-distributed (OOD) regarding graph sizes and other characteristics of training data. \cite{bevilacqua2021size} proposed to learn environment-robust graph representation by describing the mechanism of distribution shift. In addition, GLNNs aiming at the OOD scenario are almost unexplored. We refer readers who are interested in this issue to this paper \cite{hu2020open} for graph data splitting that satisfies OOD.

\subsection{Performant Parameters}
As the non-convex model, GLNNs-GNNs need careful initialization and training to obtain performant parameters. However, with the ever-complex architecture of GLNNs-GNNs and the increasing size of the data scale, searching performant parameters becomes GPU-consuming. \cite{knyazev2021parameter} proposed a model that learns knowledge about predicting performant parameters at the training stage and predicts performant parameters for unseen deep learning architectures through a single forward pass at the test stage. Such an optimization architecture exposes determining performant parameters of GLNNs-GNNs to a new direction.

\subsection{Applying GLNNs to Classify Brain Networks}
Brain network data are scarce since obtaining a network needs to scan an individual's brain. A brain network contains a small number of nodes that denote the brain's region of interest (ROI). Different brain networks are composed of the same ROIs. Neuroscientists desire to understand which regions play a decisive role when distinguishing brains having neurological disorders from healthy individuals. Hence, brain network classification needs to simplify GLNNs that have a small number of parameters, identity awareness, and high explanation. However, existing GLNNs do not meet the requirements mentioned above \cite{lanciano2020explainable}.

\subsection{Encoding Relations among Graphs}
GLNNs treat each graph as an independent individual. However, graphs can be regarded as interrelating entities. For example, relations among graphs could be established based on common substructures. Building relations among graphs and encoding them into graph representations have the potential to tackle the issue that GLNNs only embed limited structure information into graph representations \cite{wang2020gognn}.

\section{Conclusions}
GLNNs have brought a new paradigm and a large number of state-of-the-art models for the graph-level learning community. A considerable number of high-impact publications about GLNNs are badly in need of a survey with systematic taxonomy. In this survey, we briefly reviewed traditional graph-level learning methods. Then we summarized GLNNs into three categories and deeply investigated the development and contributions of each category. The reproducibility and benchmarks of GLNNs and new graph datasets were also discussed. Finally, we identified seven potential future directions to further promote the research area with fast-growing.

\clearpage
\small
\bibliographystyle{named}
\bibliography{ijcai22}

\begin{thebibliography}{}

\bibitem[\protect\citeauthoryear{Alon and Yahav}{2021}]{alon2021on}
U.~Alon and E.~Yahav.
\newblock On the bottleneck of graph neural networks and its practical
  implications.
\newblock In {\em ICLR}, 2021.

\bibitem[\protect\citeauthoryear{Arvind \bgroup \em et al.\egroup
  }{2020}]{arvind2020weisfeiler}
V.~Arvind, F.~Fuhlbr{\"u}ck, J.~K{\"o}bler, et~al.
\newblock On weisfeiler-leman invariance: subgraph counts and related graph
  properties.
\newblock {\em J Comput Syst Sci}, 113:42--59, 2020.

\bibitem[\protect\citeauthoryear{Atwood and
  Towsley}{2016}]{atwood2016diffusion}
J.~Atwood and D.~Towsley.
\newblock Diffusion-convolutional neural networks.
\newblock In {\em NeurIPS}, 2016.

\bibitem[\protect\citeauthoryear{Babai}{2016}]{babai2016graph}
L.~Babai.
\newblock Graph isomorphism in quasipolynomial time.
\newblock In {\em STOC}, pages 684--697, 2016.

\bibitem[\protect\citeauthoryear{Balcilar \bgroup \em et al.\egroup
  }{2020}]{balcilar2020analyzing}
M.~Balcilar, G.~Renton, P.~H{\'e}roux, et~al.
\newblock Analyzing the expressive power of graph neural networks in a spectral
  perspective.
\newblock In {\em ICLR}, 2020.

\bibitem[\protect\citeauthoryear{Balcilar \bgroup \em et al.\egroup
  }{2021}]{balcilar2021breaking}
M.~Balcilar, P.~H{\'e}roux, B.~Ga{\"u}z{\`e}re, et~al.
\newblock Breaking the limits of message passing graph neural networks.
\newblock In {\em ICML}, 2021.

\bibitem[\protect\citeauthoryear{Battaglia \bgroup \em et al.\egroup
  }{2018}]{battaglia2018relational}
P.~W. Battaglia, J.~B. Hamrick, V.~Bapst, et~al.
\newblock Relational inductive biases, deep learning, and graph networks.
\newblock {\em arXiv}, 2018.

\bibitem[\protect\citeauthoryear{Bevilacqua \bgroup \em et al.\egroup
  }{2021}]{bevilacqua2021size}
B.~Bevilacqua, Y.~Zhou, and B.~Ribeiro.
\newblock Size-invariant graph representations for graph classification
  extrapolations.
\newblock In {\em ICML}, 2021.

\bibitem[\protect\citeauthoryear{Bevilacqua \bgroup \em et al.\egroup
  }{2022}]{bevilacqua2022equivariant}
B.~Bevilacqua, F.~Frasca, D.~Lim, et~al.
\newblock Equivariant subgraph aggregation networks.
\newblock In {\em ICLR}, 2022.

\bibitem[\protect\citeauthoryear{Bianchi \bgroup \em et al.\egroup
  }{2020}]{bianchi2020spectral}
F.~M. Bianchi, D.~Grattarola, and C.~Alippi.
\newblock Spectral clustering with graph neural networks for graph pooling.
\newblock In {\em ICML}, pages 874--883, 2020.

\bibitem[\protect\citeauthoryear{Biggs \bgroup \em et al.\egroup
  }{1986}]{biggs1986graph}
N.~Biggs, E.~K. Lloyd, and R.~J. Wilson.
\newblock {\em Graph Theory, 1736-1936}.
\newblock OUP, 1986.

\bibitem[\protect\citeauthoryear{Bodnar \bgroup \em et al.\egroup
  }{2021}]{bodnar2021weisfeiler}
C.~Bodnar, F.~Frasca, Y.~G. Wang, et~al.
\newblock Weisfeiler and lehman go topological: Message passing simplicial
  networks.
\newblock In {\em ICML}, 2021.

\bibitem[\protect\citeauthoryear{Bondy and Hemminger}{1977}]{bondy1977graph}
J.~A. Bondy and R.~L. Hemminger.
\newblock Graph reconstruction—a survey.
\newblock {\em J Graph Theory}, 1(3), 1977.

\bibitem[\protect\citeauthoryear{Bouritsas \bgroup \em et al.\egroup
  }{2020}]{bouritsas2020improving}
G.~Bouritsas, F.~Frasca, S.~Zafeiriou, et~al.
\newblock Improving graph neural network expressivity via subgraph isomorphism
  counting.
\newblock {\em arXiv}, 2020.

\bibitem[\protect\citeauthoryear{Bruna \bgroup \em et al.\egroup
  }{2014}]{bruna2014spectral}
J.~Bruna, W.~Zaremba, A.~Szlam, et~al.
\newblock Spectral networks and locally connected networks on graphs.
\newblock In {\em ICLR}, 2014.

\bibitem[\protect\citeauthoryear{Cangea \bgroup \em et al.\egroup
  }{2018}]{cangea2018towards}
C.~Cangea, P.~Veli{\v{c}}kovi{\'c}, N.~Jovanovi{\'c}, et~al.
\newblock Towards sparse hierarchical graph classifiers.
\newblock In {\em NeurIPS}, 2018.

\bibitem[\protect\citeauthoryear{Chanussot \bgroup \em et al.\egroup
  }{2021}]{chanussot2021open}
L.~Chanussot, A.~Das, S.~Goyal, et~al.
\newblock Open catalyst 2020 (oc20) dataset and community challenges.
\newblock {\em ACS Catal.}, 11(10):6059--6072, 2021.

\bibitem[\protect\citeauthoryear{Chauhan \bgroup \em et al.\egroup
  }{2020}]{chauhan2020few}
J.~Chauhan, D.~Nathani, and M.~Kaul.
\newblock Few-shot learning on graphs via super-classes based on graph spectral
  measures.
\newblock In {\em ICLR}, 2020.

\bibitem[\protect\citeauthoryear{Chen \bgroup \em et al.\egroup
  }{2018}]{chen2018rise}
H.~Chen, O.~Engkvist, Y.~Wang, et~al.
\newblock The rise of deep learning in drug discovery.
\newblock {\em Drug Discov}, 23(6), 2018.

\bibitem[\protect\citeauthoryear{Chen \bgroup \em et al.\egroup
  }{2020a}]{chen2020convolutional}
D.~Chen, L.~Jacob, and J.~Mairal.
\newblock Convolutional kernel networks for graph-structured data.
\newblock In {\em ICLR}, 2020.

\bibitem[\protect\citeauthoryear{Chen \bgroup \em et al.\egroup
  }{2020b}]{chen2020can}
Z.~Chen, L.~Chen, S.~Villar, et~al.
\newblock Can graph neural networks count substructures?
\newblock In {\em NeurIPS}, 2020.

\bibitem[\protect\citeauthoryear{Cui \bgroup \em et al.\egroup
  }{2018}]{cui2018survey}
P.~Cui, X.~Wang, J.~Pei, et~al.
\newblock A survey on network embedding.
\newblock {\em IEEE Trans. Knowl. Data Eng.}, 2018.

\bibitem[\protect\citeauthoryear{Du \bgroup \em et al.\egroup
  }{2019}]{du2019graph}
S.~S. Du, K.~Hou, R.~R. Salakhutdinov, et~al.
\newblock Graph neural tangent kernel: Fusing graph neural networks with graph
  kernels.
\newblock In {\em NeurIPS}, volume~32, pages 5723--5733, 2019.

\bibitem[\protect\citeauthoryear{Duvenaud \bgroup \em et al.\egroup
  }{2015}]{duvenaud2015convolutional}
D.~Duvenaud, D.~Maclaurin, J.~Aguilera-Iparraguirre, et~al.
\newblock Convolutional networks on graphs for learning molecular fingerprints.
\newblock In {\em NeurIPS}, 2015.

\bibitem[\protect\citeauthoryear{Dwivedi \bgroup \em et al.\egroup
  }{2020}]{dwivedi2020benchmarkgnns}
V.~P. Dwivedi, C.~K. Joshi, T.~Laurent, et~al.
\newblock Benchmarking graph neural networks.
\newblock {\em arXiv}, 2020.

\bibitem[\protect\citeauthoryear{Errica \bgroup \em et al.\egroup
  }{2020}]{errica2019fair}
F.~Errica, M.~Podda, D.~Bacciu, et~al.
\newblock A fair comparison of graph neural networks for graph classification.
\newblock In {\em ICLR}, 2020.

\bibitem[\protect\citeauthoryear{Freitas \bgroup \em et al.\egroup
  }{2021}]{freitas2020large}
S.~Freitas, Y.~Dong, J.~Neil, et~al.
\newblock A large-scale database for graph representation learning.
\newblock In {\em NeurIPS}, 2021.

\bibitem[\protect\citeauthoryear{Gao and Ji}{2019}]{gao2019graph}
H.~Gao and S.~Ji.
\newblock Graph u-nets.
\newblock In {\em ICML}, 2019.

\bibitem[\protect\citeauthoryear{Geerts and
  Reutter}{2022}]{geerts2022expressiveness}
F.~Geerts and J.~L. Reutter.
\newblock Expressiveness and approximation properties of graph neural networks.
\newblock In {\em ICLR}, 2022.

\bibitem[\protect\citeauthoryear{Gilmer \bgroup \em et al.\egroup
  }{2017}]{gilmer2017neural}
J.~Gilmer, S.~S. Schoenholz, P.~F. Riley, et~al.
\newblock Neural message passing for quantum chemistry.
\newblock In {\em ICML}, 2017.

\bibitem[\protect\citeauthoryear{Haan \bgroup \em et al.\egroup
  }{2020}]{de2020natural}
P.~Haan, de, T.~Cohen, and M.~Welling.
\newblock Natural graph networks.
\newblock In {\em NeurIPS}, 2020.

\bibitem[\protect\citeauthoryear{Hinton \bgroup \em et al.\egroup
  }{2011}]{hinton2011transforming}
G.~E. Hinton, A.~Krizhevsky, and S.~D. Wang.
\newblock Transforming auto-encoders.
\newblock In {\em ICANN}, pages 44--51, 2011.

\bibitem[\protect\citeauthoryear{Hjelm \bgroup \em et al.\egroup
  }{2019}]{hjelm2019learning}
R.~D. Hjelm, A.~Fedorov, S.~Lavoie-Marchildon, et~al.
\newblock Learning deep representations by mutual information estimation and
  maximization.
\newblock In {\em ICLR}, 2019.

\bibitem[\protect\citeauthoryear{Hu \bgroup \em et al.\egroup
  }{2020}]{hu2020open}
W.~Hu, M.~Fey, M.~Zitnik, et~al.
\newblock Open graph benchmark: Datasets for machine learning on graphs.
\newblock In {\em NeurIPS}, 2020.

\bibitem[\protect\citeauthoryear{Huang \bgroup \em et al.\egroup
  }{2020}]{huang2020graphlime}
Q.~Huang, M.~Yamada, Y.~Tian, and other.
\newblock Graphlime: Local interpretable model explanations for graph neural
  networks.
\newblock {\em arXiv}, 2020.

\bibitem[\protect\citeauthoryear{Ivanov and
  Burnaev}{2018}]{ivanov2018anonymous}
S.~Ivanov and E.~Burnaev.
\newblock Anonymous walk embeddings.
\newblock In {\em ICML}, pages 2186--2195, 2018.

\bibitem[\protect\citeauthoryear{Ivanov \bgroup \em et al.\egroup
  }{2019}]{ivanov2019understanding}
S.~Ivanov, S.~Sviridov, and E.~Burnaev.
\newblock Understanding isomorphism bias in graph data sets.
\newblock {\em arXiv}, 2019.

\bibitem[\protect\citeauthoryear{Kang \bgroup \em et al.\egroup
  }{2012}]{kang2012fast}
U.~Kang, H.~Tong, and J.~Sun.
\newblock Fast random walk graph kernel.
\newblock In {\em SDM}, pages 828--838. SIAM, 2012.

\bibitem[\protect\citeauthoryear{Khasahmadi \bgroup \em et al.\egroup
  }{2020}]{Khasahmadi2020Memory-Based}
A.~H. Khasahmadi, K.~Hassani, P.~Moradi, et~al.
\newblock Memory-based graph networks.
\newblock In {\em ICLR}, 2020.

\bibitem[\protect\citeauthoryear{Kipf and Welling}{2017}]{kipf2016semi}
T.~N. Kipf and M.~Welling.
\newblock Semi-supervised classification with graph convolutional networks.
\newblock In {\em ICLR}, 2017.

\bibitem[\protect\citeauthoryear{Klicpera \bgroup \em et al.\egroup
  }{2021}]{klicpera2021gemnet}
J.~Klicpera, F.~Becker, and S.~G{\"u}nnemann.
\newblock Gemnet: Universal directional graph neural networks for molecules.
\newblock In {\em NeurIPS}, 2021.

\bibitem[\protect\citeauthoryear{Knyazev \bgroup \em et al.\egroup
  }{2021}]{knyazev2021parameter}
B.~Knyazev, M.~Drozdzal, G.~W. Taylor, et~al.
\newblock Parameter prediction for unseen deep architectures.
\newblock In {\em NeurIPS}, 2021.

\bibitem[\protect\citeauthoryear{Kondor \bgroup \em et al.\egroup
  }{2018}]{kondor2018covariant}
R.~Kondor, H.~T. Son, H.~Pan, et~al.
\newblock Covariant compositional networks for learning graphs.
\newblock In {\em ICLR}, 2018.

\bibitem[\protect\citeauthoryear{Kriege \bgroup \em et al.\egroup
  }{2020}]{kriege2020survey}
N.~M. Kriege, F.~D. Johansson, and C.~Morris.
\newblock A survey on graph kernels.
\newblock {\em Appl. Netw. Sci.}, 5(1):1--42, 2020.

\bibitem[\protect\citeauthoryear{Kudo \bgroup \em et al.\egroup
  }{2004}]{kudo2004application}
T.~Kudo, E.~Maeda, and Y.~Matsumoto.
\newblock An application of boosting to graph classification.
\newblock In {\em NeurIPS}, 2004.

\bibitem[\protect\citeauthoryear{Lanciano \bgroup \em et al.\egroup
  }{2020}]{lanciano2020explainable}
T.~Lanciano, F.~Bonchi, and A.~Gionis.
\newblock Explainable classification of brain networks via contrast subgraphs.
\newblock In {\em KDD}, pages 3308--3318, 2020.

\bibitem[\protect\citeauthoryear{Lee \bgroup \em et al.\egroup
  }{2018}]{lee2018graph}
J.~B. Lee, R.~Rossi, and X.~Kong.
\newblock Graph classification using structural attention.
\newblock In {\em KDD}, 2018.

\bibitem[\protect\citeauthoryear{Lee \bgroup \em et al.\egroup
  }{2019}]{lee2019self}
J.~Lee, I.~Lee, and J.~Kang.
\newblock Self-attention graph pooling.
\newblock In {\em ICML}, pages 3734--3743, 2019.

\bibitem[\protect\citeauthoryear{Lee \bgroup \em et al.\egroup
  }{2021}]{lee2021learnable}
D.~Lee, S.~Kim, S.~Lee, et~al.
\newblock Learnable structural semantic readout for graph classification.
\newblock In {\em ICDM}, 2021.

\bibitem[\protect\citeauthoryear{Lei \bgroup \em et al.\egroup
  }{2017}]{lei2017deriving}
T.~Lei, W.~Jin, R.~Barzilay, et~al.
\newblock Deriving neural architectures from sequence and graph kernels.
\newblock In {\em ICML}, 2017.

\bibitem[\protect\citeauthoryear{Long \bgroup \em et al.\egroup
  }{2021}]{long2021theoretically}
Q.~Long, Y.~Jin, Y.~Wu, et~al.
\newblock Theoretically improving graph neural networks via anonymous walk
  graph kernels.
\newblock In {\em WWW}, pages 1204--1214, 2021.

\bibitem[\protect\citeauthoryear{Luo \bgroup \em et al.\egroup
  }{2020}]{luo2020parameterized}
D.~Luo, W.~Cheng, D.~Xu, et~al.
\newblock Parameterized explainer for graph neural network.
\newblock In {\em NeurIPS}, 2020.

\bibitem[\protect\citeauthoryear{Ma \bgroup \em et al.\egroup
  }{2019}]{ma2019graph}
Y.~Ma, S.~Wang, C.~C. Aggarwal, et~al.
\newblock Graph convolutional networks with eigenpooling.
\newblock In {\em KDD}, 2019.

\bibitem[\protect\citeauthoryear{Ma \bgroup \em et al.\egroup
  }{2020}]{ma2020adaptive}
N.~Ma, J.~Bu, J.~Yang, et~al.
\newblock Adaptive-step graph meta-learner for few-shot graph classification.
\newblock In {\em CIKM}, 2020.

\bibitem[\protect\citeauthoryear{Ma \bgroup \em et al.\egroup
  }{2021}]{ma2021comprehensive}
X.~Ma, J.~Wu, S.~Xue, et~al.
\newblock A comprehensive survey on graph anomaly detection with deep learning.
\newblock {\em IEEE Trans. Knowl. Data Eng.}, 2021.

\bibitem[\protect\citeauthoryear{Maron \bgroup \em et al.\egroup
  }{2019}]{maron2019provably}
H.~Maron, H.~Ben-Hamu, H.~Serviansky, et~al.
\newblock Provably powerful graph networks.
\newblock In {\em NeurIPS}, 2019.

\bibitem[\protect\citeauthoryear{McKay and Piperno}{2014}]{mckay2014practical}
B.~D. McKay and A.~Piperno.
\newblock Practical graph isomorphism, ii.
\newblock {\em J Symb Comput}, 60:94--112, 2014.

\bibitem[\protect\citeauthoryear{Medsker and Jain}{2001}]{medsker2001recurrent}
L.~R. Medsker and L.~Jain.
\newblock Recurrent neural networks.
\newblock {\em Design and Applications}, 5:64--67, 2001.

\bibitem[\protect\citeauthoryear{Mesquita \bgroup \em et al.\egroup
  }{2020}]{mesquita2020rethinking}
D.~Mesquita, A.~H. Souza, and S.~Kaski.
\newblock Rethinking pooling in graph neural networks.
\newblock In {\em NeurIPS}, 2020.

\bibitem[\protect\citeauthoryear{Morgan}{1965}]{morgan1965generation}
H.~L. Morgan.
\newblock The generation of a unique machine description for chemical
  structures-a technique developed at chemical abstracts service.
\newblock {\em J. chem. doc.}, 5(2):107--113, 1965.

\bibitem[\protect\citeauthoryear{Morris \bgroup \em et al.\egroup
  }{2019}]{morris2019weisfeiler}
C.~Morris, M.~Ritzert, M.~Fey, et~al.
\newblock Weisfeiler and leman go neural: Higher-order graph neural networks.
\newblock In {\em AAAI}, pages 4602--4609, 2019.

\bibitem[\protect\citeauthoryear{Morris \bgroup \em et al.\egroup
  }{2020}]{morris2020tudataset}
C.~Morris, N.~M. Kriege, F.~Bause, et~al.
\newblock Benchmark data sets for graph kernels.
\newblock \url{https://chrsmrrs.github.io/datasets/}, 2020.

\bibitem[\protect\citeauthoryear{Murphy \bgroup \em et al.\egroup
  }{2019}]{murphy2019relational}
R.~Murphy, B.~Srinivasan, V.~Rao, et~al.
\newblock Relational pooling for graph representations.
\newblock In {\em ICML}, 2019.

\bibitem[\protect\citeauthoryear{Narayanan \bgroup \em et al.\egroup
  }{2017}]{narayanan2017graph2vec}
A.~Narayanan, M.~Chandramohan, R.~Venkatesan, et~al.
\newblock graph2vec: Learning distributed representations of graphs.
\newblock {\em CoRR}, abs/1707.05005, 2017.

\bibitem[\protect\citeauthoryear{Niepert \bgroup \em et al.\egroup
  }{2016}]{niepert2016learning}
M.~Niepert, M.~Ahmed, and K.~Kutzkov.
\newblock Learning convolutional neural networks for graphs.
\newblock In {\em ICML}, pages 2014--2023, 2016.

\bibitem[\protect\citeauthoryear{Nikolentzos \bgroup \em et al.\egroup
  }{2021}]{nikolentzos2021graph}
G.~Nikolentzos, G.~Siglidis, and M.~Vazirgiannis.
\newblock Graph kernels: A survey.
\newblock {\em J. Artif. Intell.}, 2021.

\bibitem[\protect\citeauthoryear{Pope \bgroup \em et al.\egroup
  }{2019}]{pope2019explainability}
P.~E. Pope, S.~Kolouri, M.~Rostami, et~al.
\newblock Explainability methods for graph convolutional neural networks.
\newblock In {\em CVPR}, pages 10772--10781, 2019.

\bibitem[\protect\citeauthoryear{Qiu \bgroup \em et al.\egroup
  }{2020}]{qiu2020gcc}
J.~Qiu, Q.~Chen, Y.~Dong, et~al.
\newblock Gcc: Graph contrastive coding for graph neural network pre-training.
\newblock In {\em KDD}, pages 1150--1160, 2020.

\bibitem[\protect\citeauthoryear{Ralaivola \bgroup \em et al.\egroup
  }{2005}]{ralaivola2005graph}
L.~Ralaivola, S.~J. Swamidass, H.~Saigo, et~al.
\newblock Graph kernels for chemical informatics.
\newblock {\em Neural Netw}, 18(8):1093--1110, 2005.

\bibitem[\protect\citeauthoryear{Ranjan \bgroup \em et al.\egroup
  }{2020}]{ranjan2020asap}
E.~Ranjan, S.~Sanyal, and P.~Talukdar.
\newblock Asap: Adaptive structure aware pooling for learning hierarchical
  graph representations.
\newblock In {\em AAAI}, pages 5470--5477, 2020.

\bibitem[\protect\citeauthoryear{Sato}{2020}]{sato2020survey}
R.~Sato.
\newblock A survey on the expressive power of graph neural networks.
\newblock {\em arXiv}, 2020.

\bibitem[\protect\citeauthoryear{Satorras \bgroup \em et al.\egroup
  }{2021}]{satorras2021n}
G.~Satorras, E.~Hoogeboom, and M.~Welling.
\newblock E(n) equivariant graph neural networks.
\newblock In {\em ICML}, 2021.

\bibitem[\protect\citeauthoryear{Sawlani \bgroup \em et al.\egroup
  }{2021}]{sawlani2021fast}
S.~Sawlani, L.~Zhao, and L.~Akoglu.
\newblock Fast attributed graph embedding via density of states.
\newblock In {\em ICDM}, 2021.

\bibitem[\protect\citeauthoryear{Shervashidze \bgroup \em et al.\egroup
  }{2009}]{shervashidze2009efficient}
N.~Shervashidze, S.~Vishwanathan, T.~Petri, et~al.
\newblock Efficient graphlet kernels for large graph comparison.
\newblock In {\em AISTATS}, pages 488--495, 2009.

\bibitem[\protect\citeauthoryear{Shervashidze \bgroup \em et al.\egroup
  }{2011}]{shervashidze2011weisfeiler}
N.~Shervashidze, P.~Schweitzer, E.~J. Van~Leeuwen, et~al.
\newblock Weisfeiler-lehman graph kernels.
\newblock {\em J Mach Learn Res}, 12(9), 2011.

\bibitem[\protect\citeauthoryear{Simonovsky and
  Komodakis}{2017}]{simonovsky2017dynamic}
M.~Simonovsky and N.~Komodakis.
\newblock Dynamic edge-conditioned filters in convolutional neural networks on
  graphs.
\newblock In {\em CVPR}, pages 3693--3702, 2017.

\bibitem[\protect\citeauthoryear{Su \bgroup \em et al.\egroup
  }{2021}]{su2021comprehensive}
X.~Su, S.~Xue, F.~Liu, et~al.
\newblock A comprehensive survey on community detection with deep learning.
\newblock {\em IEEE Trans. Neural Netw. Learn. Syst.}, 2021.

\bibitem[\protect\citeauthoryear{Sun \bgroup \em et al.\egroup
  }{2019}]{sun2019infograph}
F.~Sun, J.~Hoffmann, V.~Verma, et~al.
\newblock Infograph: Unsupervised and semi-supervised graph-level
  representation learning via mutual information maximization.
\newblock In {\em ICLR}, 2019.

\bibitem[\protect\citeauthoryear{Thoma \bgroup \em et al.\egroup
  }{2010}]{thoma2010discriminative}
M.~Thoma, H.~Cheng, A.~Gretton, et~al.
\newblock Discriminative frequent subgraph mining with optimality guarantees.
\newblock {\em Stat Anal Data Min}, 3(5):302--318, 2010.

\bibitem[\protect\citeauthoryear{Togninalli \bgroup \em et al.\egroup
  }{2019}]{togninalli2019wasserstein}
M.~Togninalli, E.~Ghisu, F.~Llinares-L{\'o}pez, et~al.
\newblock Wasserstein weisfeiler-lehman graph kernels.
\newblock In {\em NeurIPS}, 2019.

\bibitem[\protect\citeauthoryear{Verma and Zhang}{2018}]{verma2018graph}
S.~Verma and Z.~Zhang.
\newblock Graph capsule convolutional neural networks.
\newblock In {\em ICML\&IJCAI workshop}, 2018.

\bibitem[\protect\citeauthoryear{Wang \bgroup \em et al.\egroup
  }{2017}]{wang2017structural}
S.~Wang, L.~He, B.~Cao, et~al.
\newblock Structural deep brain network mining.
\newblock In {\em KDD}, pages 475--484, 2017.

\bibitem[\protect\citeauthoryear{Wang \bgroup \em et al.\egroup
  }{2019}]{wang2019dgl}
M.~Wang, D.~Zheng, Z.~Ye, et~al.
\newblock Deep graph library: A graph-centric, highly-performant package for
  graph neural networks.
\newblock In {\em ICLR Workshop}, 2019.

\bibitem[\protect\citeauthoryear{Wang \bgroup \em et al.\egroup
  }{2020}]{wang2020gognn}
H.~Wang, D.~Lian, Y.~Zhang, L.~Qin, and X.~Lin.
\newblock Gognn: Graph of graphs neural network for predicting structured
  entity interactions.
\newblock In {\em IJCAI}, 2020.

\bibitem[\protect\citeauthoryear{Wei \bgroup \em et al.\egroup
  }{2021}]{wei2021pooling}
L.~Wei, H.~Zhao, Q.~Yao, et~al.
\newblock Pooling architecture search for graph classification.
\newblock In {\em CIKM}, 2021.

\bibitem[\protect\citeauthoryear{Weisfeiler and
  Leman}{1968}]{weisfeiler1968reduction}
B.~Weisfeiler and A.~Leman.
\newblock The reduction of a graph to canonical form and the algebra which
  appears therein.
\newblock {\em NTI, Series}, 2(9):12--16, 1968.

\bibitem[\protect\citeauthoryear{Wijesinghe and Wang}{2022}]{wijesinghe2022a}
A.~Wijesinghe and Q.~Wang.
\newblock A new perspective on ''how graph neural networks go beyond
  weisfeiler-lehman?''.
\newblock In {\em ICLR}, 2022.

\bibitem[\protect\citeauthoryear{Wu \bgroup \em et al.\egroup
  }{2020}]{wu2020comprehensive}
Z.~Wu, S.~Pan, F.~Chen, et~al.
\newblock A comprehensive survey on graph neural networks.
\newblock {\em IEEE Trans Neural Netw Learn Syst}, 32(1):4--24, 2020.

\bibitem[\protect\citeauthoryear{Xu \bgroup \em et al.\egroup
  }{2018}]{xu2018powerful}
K.~Xu, W.~Hu, J.~Leskovec, et~al.
\newblock How powerful are graph neural networks?
\newblock In {\em NeurIPS}, 2018.

\bibitem[\protect\citeauthoryear{Xu \bgroup \em et al.\egroup
  }{2021}]{xu2020neural}
K.~Xu, M.~Zhang, J.~Li, et~al.
\newblock How neural networks extrapolate: From feedforward to graph neural
  networks.
\newblock In {\em ICLR}, 2021.

\bibitem[\protect\citeauthoryear{Yehudai \bgroup \em et al.\egroup
  }{2021}]{yehudai2021local}
G.~Yehudai, E.~Fetaya, E.~Meirom, et~al.
\newblock From local structures to size generalization in graph neural
  networks.
\newblock In {\em ICML}, pages 11975--11986, 2021.

\bibitem[\protect\citeauthoryear{Ying \bgroup \em et al.\egroup
  }{2018}]{ying2018hierarchical}
R.~Ying, J.~You, C.~Morris, et~al.
\newblock Hierarchical graph representation learning with differentiable
  pooling.
\newblock In {\em NeurIPS}, 2018.

\bibitem[\protect\citeauthoryear{Ying \bgroup \em et al.\egroup
  }{2019}]{ying2019gnnexplainer}
R.~Ying, D.~Bourgeois, J.~You, et~al.
\newblock Gnnexplainer: Generating explanations for graph neural networks.
\newblock In {\em NeurIPS}, 2019.

\bibitem[\protect\citeauthoryear{You \bgroup \em et al.\egroup
  }{2018}]{you2018graphrnn}
J.~You, R.~Ying, X.~Ren, et~al.
\newblock Graphrnn: Generating realistic graphs with deep auto-regressive
  models.
\newblock In {\em ICML}, pages 5708--5717, 2018.

\bibitem[\protect\citeauthoryear{You \bgroup \em et al.\egroup
  }{2020}]{you2020graph}
Y.~You, T.~Chen, Y.~Sui, et~al.
\newblock Graph contrastive learning with augmentations.
\newblock In {\em NeurIPS}, 2020.

\bibitem[\protect\citeauthoryear{Yuan and Ji}{2020}]{yuan2020StructPool}
H.~Yuan and S.~Ji.
\newblock Structpool: Structured graph pooling via conditional random fields.
\newblock In {\em ICLR}, 2020.

\bibitem[\protect\citeauthoryear{Yuan \bgroup \em et al.\egroup
  }{2020a}]{yuan2020xgnn}
H.~Yuan, J.~Tang, X.~Hu, et~al.
\newblock Xgnn:towards model-level explanations of graph neural networks.
\newblock In {\em KDD}, 2020.

\bibitem[\protect\citeauthoryear{Yuan \bgroup \em et al.\egroup
  }{2020b}]{yuan2020explainability}
H.~Yuan, H.~Yu, S.~Gui, et~al.
\newblock Explainability in graph neural networks: A taxonomic survey.
\newblock {\em arXiv}, 2020.

\bibitem[\protect\citeauthoryear{Zhang and Chen}{2018}]{xinyi2018capsule}
X.~Zhang and L.~Chen.
\newblock Capsule graph neural network.
\newblock In {\em ICLR}, 2018.

\bibitem[\protect\citeauthoryear{Zhang \bgroup \em et al.\egroup
  }{2018}]{zhang2018end}
M.~Zhang, Z.~Cui, M.~Neumann, et~al.
\newblock An end-to-end deep learning architecture for graph classification.
\newblock In {\em AAAI}, 2018.

\bibitem[\protect\citeauthoryear{Zhang \bgroup \em et al.\egroup
  }{2020}]{zhang2020deep}
Z.~Zhang, P.~Cui, and W.~Zhu.
\newblock Deep learning on graphs: A survey.
\newblock {\em IEEE Trans. Knowl. Data Eng.}, 34(1), 2020.

\bibitem[\protect\citeauthoryear{Zheng \bgroup \em et al.\egroup
  }{2021}]{zheng2021framelets}
X.~Zheng, B.~Zhou, J.~Gao, et~al.
\newblock How framelets enhance graph neural networks.
\newblock {\em ICML}, 2021.

\end{thebibliography}

\end{document}